\documentclass[letterpaper, 10 pt, conference]{ieeeconf}
\IEEEoverridecommandlockouts 

\usepackage{microtype}

\usepackage{hyperref}
\hypersetup{colorlinks,linkcolor={green!50!black},citecolor={green!50!black},urlcolor={blue!80!black}}
\makeatletter \let\NAT@parse\undefined \makeatother
\usepackage[sort,compress]{cite}
\usepackage{graphicx} 
\usepackage{amsfonts}
\usepackage{amsmath,soul}
\usepackage{color}
\usepackage[font=small]{subcaption}
\usepackage{balance}
\usepackage[font=small]{caption}
\usepackage[linesnumbered,ruled,vlined]{algorithm2e}
\usepackage{multirow}

\DeclareMathOperator*{\argmin}{\arg\!\min}
\DeclareMathOperator*{\argmax}{\arg\!\max}
\usepackage{tabulary}
\newcolumntype{K}[1]{>{\centering\arraybackslash}p{#1}}

\title{\LARGE \bf Motion Planning Among Dynamic, Decision-Making Agents\\ with Deep Reinforcement Learning}

\author{Michael Everett$^\ddag$, Yu Fan Chen$^\dag$, and Jonathan P.\ How$^\ddag$
  \thanks{$^\ddag$Aerospace Controls Laboratory,
    Massachusetts Institute of Technology, 77 Massachusetts Ave.,
    Cambridge, MA, USA. {\tt\footnotesize \{mfe, jhow\}@mit.edu}}%
   \thanks{$^\dag$Oculus Research, Redmond, WA, USA {\tt\footnotesize steven.chen2@oculus.com }}
}

\usepackage[svgnames]{xcolor} \definecolor{DarkGreen}{rgb}{0,0.5,0}
\definecolor{DarkRed}{rgb}{0.75,0,0}

\usepackage[authormarkuptext=name,addedmarkup=bf,authormarkupposition=left]{changes}
\definechangesauthor[name={J.~H.}, color={blue}]{jh}
\definechangesauthor[name={M.~E.}, color={red}]{me}
\setremarkmarkup{(#2)}

\usepackage{tikz,mathtools}
\usepackage[capitalize]{cleveref}
\crefformat{equation}{(#2#1#3)}
\Crefformat{equation}{Equation~(#2#1#3)}
\Crefname{equation}{Equation}{Equations}

\usetikzlibrary{shapes,positioning,automata,arrows,fit,backgrounds,calc}
\tikzstyle{block} = [draw, fill=blue!20, rectangle,minimum height=1em,
minimum width=2em] \tikzstyle{sum} = [draw, fill=blue!20, circle, node
distance=1cm] \tikzstyle{input} = [coordinate] \tikzstyle{output} =
[coordinate] \tikzstyle{pinstyle} = [pin edge={to-,thin,black}]
\usetikzlibrary{trees} \usetikzlibrary{decorations.pathmorphing}
\usetikzlibrary{decorations.markings}
\definecolor{darkgreen}{rgb}{0,0.5,0}
\definecolor{darkred}{rgb}{220,20,60}

\makeatletter
\renewcommand\paragraph{\@startsection{subsubsection}{4}{\z@}%
{0.25ex \@plus.5ex \@minus.2ex}%
{-.15em}%
{\normalfont\normalsize\itshape}}
\makeatother

\begin{document}

\maketitle
\thispagestyle{empty} \pagestyle{empty}

\begin{abstract}
Robots that navigate among pedestrians use collision avoidance algorithms to enable safe and efficient operation. 
Recent works present deep reinforcement learning as a framework to model the complex interactions and cooperation.
However, they are implemented using key assumptions about other agents' behavior that deviate from reality as the number of agents in the environment increases.
This work extends our previous approach to develop an algorithm that learns collision avoidance among a variety of types of dynamic agents without assuming they follow any particular behavior rules.
This work also introduces a strategy using LSTM that enables the algorithm to use observations of an arbitrary number of other agents, instead of previous methods that have a fixed observation size. 
The proposed algorithm outperforms our previous approach in simulation as the number of agents increases, and the algorithm is demonstrated on a fully autonomous robotic vehicle traveling at human walking speed, without the use of a 3D Lidar.
\end{abstract}

\section{Introduction} \label{sec:intro}

Robots that navigate among pedestrians will observe many human behaviors, such as cooperation or obliviousness.
Not only are pedestrians moving obstacles, but they are constantly making decisions that a robot can only partially observe.
This work addresses the collision avoidance problem of an agent operating in a world of other decision-making agents, particularly considering the robot-pedestrian domain.
A fundamental question in decentralized collision avoidance algorithms is: what does the agent know and assume about other agents' belief states, policies, and intents?
Without communication between agents, these properties are not directly measurable, but it is possible that they can be inferred.

The assumptions an agent makes about the behavior of other agents affects how it decides which action to take.
In the simplest case, agents assume other agents are static, and re-plan quickly enough to avoid collisions.
Another approach assumes other agents are dynamic obstacles, but with a constant velocity~\cite{van_den_berg_reciprocal_2008}.
Further, agents can assume other agents are decision-makers, whose velocities may change at any moment according to known or unknown policies (decision rules).
Even if the robot knew the pedestrians' decision rule, because the other agents' intents are unknown (e.g.\ goal destination), it is impossible to perfectly predict how other non-communicating decision-making agents (e.g.\ pedestrians) will respond to an agent's decisions. Thus, instead of trying to explicitly predict other agents' behaviors, recent approaches have used reinforcement learning (RL) to model the complex interactions and cooperation among agents~\cite{chen_decentralized_2017,Chen17_IROS,long2017deep,long2017towards,li2017role,qi2018intent}.

Although learning-based methods have been shown to perform well in this domain, existing approaches make subtle assumptions about other agents such as homogeneity~\cite{long2017towards} or a specific motion model over short timescales~\cite{chen_decentralized_2017,Chen17_IROS}.
In this work, we extend our previous algorithms~\cite{chen_decentralized_2017,Chen17_IROS} to learn a collision avoidance policy without assuming that other agents follow any particular behavior model.

\begin{figure}[t]
	\centering
	\includegraphics [trim=0 0 0 0, clip, angle=0, width=0.75\columnwidth,
	keepaspectratio]{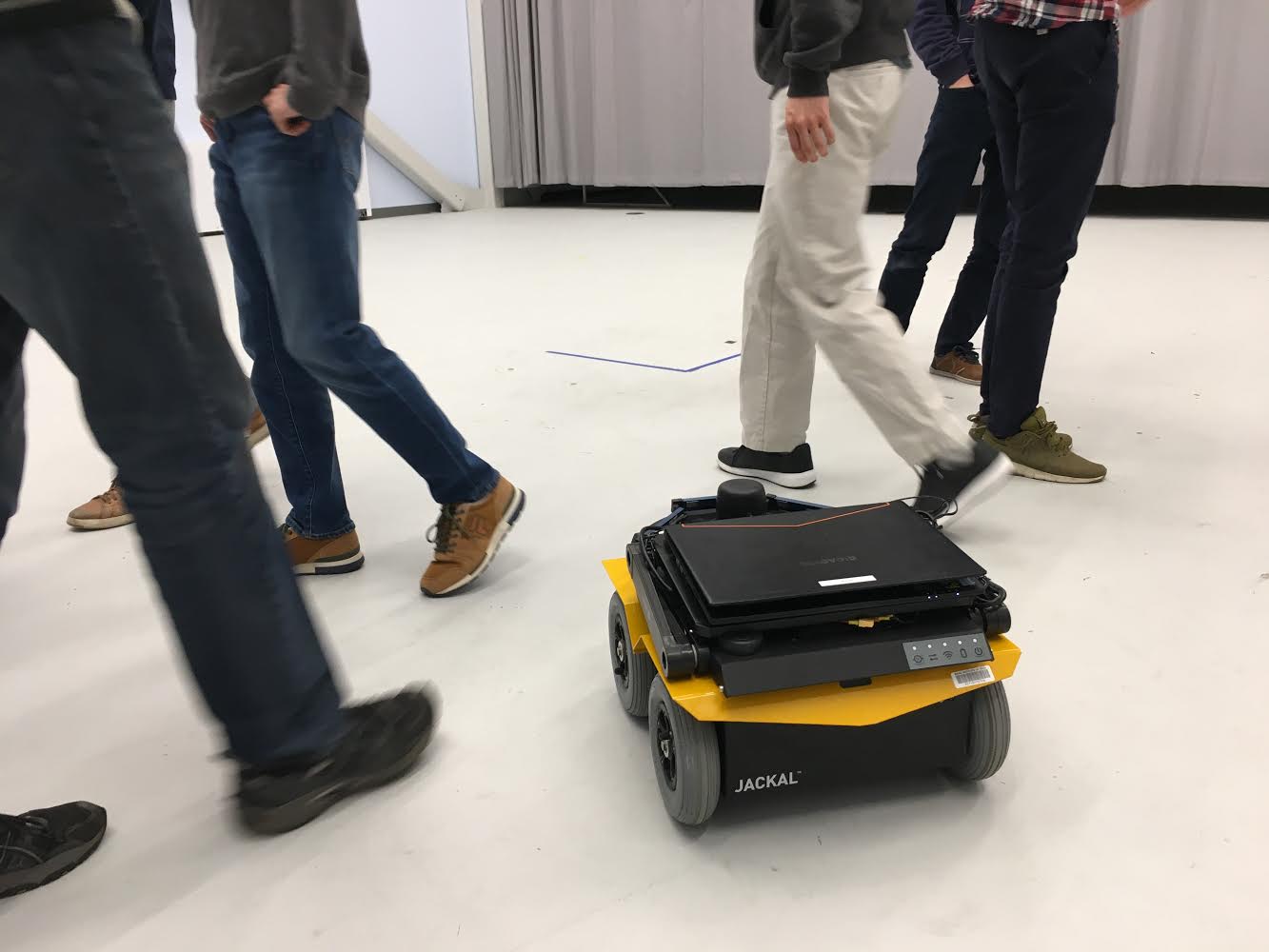}
	\caption{A robot navigates among pedestrians. Robots use onboard sensors to perceive the environment and run collsion avoidance algorithms to maintain safe and efficient operation.} 
	\label{fig:rover} 
\end{figure}

Another key challenge in collision avoidance is that the number of other agents in the environment varies, however the typical feedforward neural networks used in this domain require a fixed-dimension input.
Existing strategies define a maximum number of agents that the network can observe, or use raw sensor data as the input.
This work instead uses an idea from Natural Language Processing~\cite{sutskever2014sequence,cho2014learning} to encode the varying size state of the world (e.g. positions of other agents) into a fixed-length vector, using long short-term memory (LSTM)~\cite{hochreiter1997long} cells at the network input.
This enables the algorithm to make decisions based on an arbitrary number of other agents in the robot's vicinity.

The main contributions of this work are 
(i) an extension to our collision avoidance algorithm that does not assume the behavior of other decision-making agents,
(ii) a strategy that enables the algorithm to use observations of an arbitrary number of other agents
(iii) simulation results demonstrating the benefits of our new framework, and
(iv) demonstration of the algorithm on a robot among pedestrians, without the use of a 3D Lidar.
The software has been released as an open-source ROS package 
\texttt{cadrl\_ros}\footnote{\url{https://github.com/mfe7/cadrl_ros}}.

\section{Background}\label{sec:prob_formulation}
\subsection{Related Work}

The problem of decentralized collision avoidance among non-communicating, dynamic agents can be broadly classified into reaction-based methods and trajectory-based methods - many of which are non-learning based~\cite{chen_decentralized_2017}.
Reaction-based methods use one-step interaction rules based on geometry or physics to ensure collision avoidance, but are often short-sighted in time.
Trajectory-based methods compute plans on longer timescale to produce smoother paths, but are often computationally expensive or require knowledge of unobservable states.
Instead, some of our previous works on collision avoidance use deep reinforcement learning~\cite{chen_decentralized_2017,Chen17_IROS} to learn a value function that encodes the expected time for an agent to reach its goal from a given state.
The expensive operation of modeling the complex interactions is learned in an offline training step, whereas the learned policy can be queried quickly online, combining the benefits of both classes of methods.
This algorithm was demonstrated on a robot navigating autonomously among pedestrians at human walking speed in a wide variety of indoor and outdoor environments.
Cooperation is embedded in the learned value function, and the algorithm compares possible actions by querying the value of future states after an arbitrary forward propagation of other agents.

Other deep RL approaches~\cite{long2017towards,tai2017virtual,tai2017socially} learn to select actions directly from raw sensor readings (either 2D laserscans or images) with end-to-end training.
The raw sensor approach has the advantage that both static and dynamic obstacles (including walls) can be fed into the network with a single framework.
However in real environments, it is useful to extract an agent-level representation of the world from multiple sensors (e.g. cameras and Lidar).
For example, a trash can and stationary person may look similar in a raw laserscan return, but the person has the potential to move at any moment, and the trash can will not become uncomfortable if the robot gets too close.
This agent-level understanding of the world therefore has important implications for the robot's motion plans that are not captured in the agent-free (end-to-end) framework.

To address the challenge of a variable number of agents in the environment, one solution is to define a maximum number of agents that the network can handle, and pad the observation space if there are actually fewer agents in the environment.
This maximum number of agents is limited by the increased number of network parameters (and therefore training time) as more agents' states are added.
Another approach, using raw sensor inputs, maintains a fixed size input, but still has the same limitations.
The approach in~\cite{gupta2017cognitive} is to learn to develop an overhead map from a sequence of onboard camera views, while also learning to plan in the generated overhead map space, which was shown to work in static environments.
For dynamic environments, we do not know of a method to learn from observations of an arbitrary number of agents, that can also leverage the recent advances in multi-sensor semantic labeling applied on a agent-by-agent basis.

\subsection{Collision Avoidance with Deep RL (CADRL)}
The multiagent collision avoidance problem can be formulated as a sequential decision making problem in a reinforcement learning framework~\cite{chen_decentralized_2017,Chen17_IROS}.
Denote the agent's state, $\mathbf{s}_t$, its action, $\mathbf{u}_t$, and the state of another agent, $\tilde{\mathbf{s}_t}$.
The state vector is composed of an observable and unobservable (hidden) portion, $\mathbf{s}_t = [\mathbf{s}_t^o, \, \mathbf{s}_t^h]$.
In the global frame, observable states are the agent's position, velocity, and radius, $\mathbf{s}^o = [p_x, \, p_y, \, v_x, \, v_y, \, r] \in \mathbb{R}^{5}$, and unobservable states are the goal position, preferred speed, and orientation\footnote{Other agents' positions and velocities are straightforward to estimate with a 2D Lidar, unlike human body orientation}, $\mathbf{s}^h = [p_{gx}, \, p_{gy}, \, v_{pref}, \, \psi] \in \mathbb{R}^{4}$.
The action is a speed and heading angle, $\mathbf{u}_t = [v_t, \, \psi_t] \in \mathbb{R}^{2}$.
A policy, $\pi: \left( \mathbf{s}_t, \, \tilde{\mathbf{s}_t}^o \right) \mapsto \mathbf{u}_t$, is developed with the objective of minimizing expected time to goal $\mathbb{E}[t_g]$ while avoiding collision with other agents, 
\begin{align}
\argmin_{\pi\left(\mathbf{s}, \, \tilde{\mathbf{s}}^{o}\right)} \quad &\mathbb{E}  \left[t_g | \mathbf{s}_0, \, \tilde{\mathbf{s}}^o_0,  \, \pi \right] \label{eqn:cost} \\ 
s.t. \quad & ||\mathbf{p}_t - \tilde{\mathbf{p}}_t||_2 \geq r + \tilde{r} \qquad \forall t
		\label{eqn:con_collision} \\ 
	 \quad & \mathbf{p}_{t_g} = \mathbf{p}_g \label{eqn:con_reach_goal} \\
	 \quad & \mathbf{p}_t = \mathbf{p}_{t-1} + \Delta t \cdot \pi( \mathbf{s}_{t-1}, \, \tilde{\mathbf{s}}^o_{t-1}) \nonumber \\
	 \quad & \tilde{\mathbf{p}}_t = \tilde{\mathbf{p}}_{t-1} + \Delta t \cdot \pi( \tilde{\mathbf{s}}_{t-1}, \, \mathbf{s}^o_{t-1}), 
	 	\label{eqn:con_kinematics}
\end{align}
where \cref{eqn:con_collision} is the collision avoidance constraint, \cref{eqn:con_reach_goal} is the goal constraint, \cref{eqn:con_kinematics} is the agents' kinematics, 
and the expectation in \cref{eqn:cost} is with respect to the other agent's unobservable states (intents) and policy. 
An RL framework can be used to solve for the policy, by considering an agent's joint configuration with its neighbor, $\mathbf{s}^{jn} = \left[ \mathbf{s}, \; \tilde{\mathbf{s}}^o \right]$.
The agent is penalized for colliding with others, and rewarded for reaching its goal position, as described by a reward function, $R_{col}(\mathbf{s}^{jn}, \, \mathbf{u})$.

Previous approaches~\cite{chen_decentralized_2017,Chen17_IROS} solved this RL problem by learning an approximation to the optimal value function, $V^*(\mathbf{s}_t^{jn})$, which encodes an estimate of the expected time to goal for a particular joint configuration state.
But, a value function of the current state can not be directly implemented as a policy.
For RL problems where the subsequent state is a known function of current state and action (e.g. chess), the optimal policy, $\pi^*(\mathbf{s}^{jn}_t)$ can be generated from $V^*(\mathbf{s}_t^{jn})$, according to:
\begin{align}
& \pi^*(\mathbf{s}^{jn}_{t}) = \argmax_{\mathbf{u}} R_{col}(\mathbf{s}_{t}, \mathbf{u}) + \nonumber \\ 
& \qquad \quad \gamma^{\Delta t \cdot v_{pref}}\int_{\mathbf{s}_{t+1}^{jn}}P(\mathbf{s}^{jn}_{t+1}|\mathbf{s}^{jn}_{t}, \mathbf{u}) V^*(\mathbf{s}_{t+1}^{jn})d\mathbf{s}_{t+1}^{jn}. \label{eqn:optimal_policy}
\end{align}
This rule predicts the next state, $\mathbf{s}^{jn}_{t+1}$, from the current state, $\mathbf{s}^{jn}_{t}$, for each potential action, $\mathbf{u}$ (potentially a stochastic process), and selects $\mathbf{u}$ that leads to the state with highest value, $V^*(\mathbf{s}_{t+1}^{jn})$.

However in the collision avoidance domain, other agents' policies and intents are unknown, which means the state-transition dynamics, $P(\mathbf{s}^{jn}_{t+1}|\mathbf{s}^{jn}_{t}, \mathbf{u})$, are also unknown.
Previous approaches avoid the integral in~\cref{eqn:optimal_policy}, by assuming that other agents continue their current velocities, $\hat{\mathbf{v}}_t$, for a duration $\Delta t$, meaning the policy can be extracted from the value function
\begin{align}
\hat{\mathbf{s}}^{jn}_{t+1,\mathbf{u}} \leftarrow [\text{propagate}(\mathbf{s}_{t}, \Delta t \cdot \mathbf{u}), \text{propagate}(\tilde{\mathbf{s}}^{o}_{t}, \Delta t \cdot \hat{\mathbf{v}}_t)] \\
\pi_{CADRL}^*(\mathbf{s}^{jn}_{t}) = \argmax_{\mathbf{u}} R_{col}(\mathbf{s}_{t}, \mathbf{u}) + \gamma^{\Delta t \cdot v_{pref}} V^{*}(\hat{\mathbf{s}}^{jn}_{t+1,\mathbf{u}})\label{eqn:cadrl_policy}
\end{align}

The introduction of parameter $\Delta t$ leads to a difficult trade-off.
Due to the the approximation of the value function in a deep neural network (DNN), a sufficiently large $\Delta t$ is required such that each propagated $\mathbf{s}_{t+1,\mathbf{u}}^{jn}$ is far enough apart, which ensures $V^*(\mathbf{s}_{t+1,\mathbf{u}}^{jn})$ is not dominated by numerical noise in the network.
The implication of large $\Delta t$ is that agents are assumed to follow a constant velocity for a significant amount of time, which neglects the effects of cooperation/reactions to an agent's decisions.
As the number of agents in the environment increases, this constant velocity assumption is less likely to be valid.
Agents do not actually reach their propagated states because of the multi-agent interactions.

In addition to not capturing decision-making behavior of other agents, our experiments suggest that $\Delta t$ is a crucial parameter to ensure convergence while training the DNNs in the previous algorithms.
If $\Delta t$ is set too small or large, the training does not converge.
A value of $\Delta t = 1$ sec was experimentally determined to enable convergence, though this number does not have much theoretical rationale.
The challenge of choosing $\Delta t$ motivated the use of a different RL framework.

\subsection{Policy-Based Learning} \label{sec:background:policy}
Therefore, this work considers RL frameworks which generate a policy that an agent can execute without any arbitrary assumptions about state transition dynamics.
A recent actor-critic algorithm called A3C~\cite{mnih2016asynchronous} uses a single DNN to approximate both the value (critic) and policy (actor) functions, and is trained with two loss terms
\begin{align}
& f_{v} = (R_t - V(\mathbf{s}^{jn}_t))^2 \label{eqn:a3c_value}, \\
& f_{\pi} = \log \pi(\textbf{u}_t|\textbf{s}_t^{jn}) (R_t - V(\textbf{s}_t^{jn})) + \beta \cdot H(\pi(\textbf{s}_t^{jn})), \label{eqn:a3c_policy}
\end{align}
where \cref{eqn:a3c_value} trains the network's value output to match the future discounted reward estimate, $R_t = \sum_{i=0}^{k-1} \gamma^i r_{t+i} + \gamma^{k}V(\textbf{s}_{t+k}^{jn})$, over the next $k$ steps, just as in CADRL.
For the policy output in \cref{eqn:a3c_policy}, the first term penalizes actions which have high probability of occurring ($\log\pi$) that lead to a lower return than predicted by the value function $(R-V)$, and the second term encourages exploration by penalizing $\pi$'s entropy with tunable constant $\beta$.

In A3C, many threads of an agent interacting with an environment are simulated in parallel, and a policy is trained based on an intelligent fusion of all the agents' experiences.
The algorithm was shown to learn a policy that achieves super-human performance on many video games.
Its implementation was modified by~\cite{babaeizadeh2017ga3c} to efficiently use GPUs to maximize the number of training experiences processed per second - the so-called GA3C learns an order of magnitude faster than A3C in many cases.
As DNNs can be efficiently trained and evaluated in batches on a GPU, a main contribution of GA3C is the use of queues for training experiences and action predictions, so that the GPU always has a batch of information to process.
This requires small modifications, related to the lag between experience and training induced by queuing, to the learning~\cref{eqn:a3c_value,eqn:a3c_policy}.
Our work builds on open-source GA3C implementations~\cite{babaeizadeh2017ga3c,omidshafiei2017crossmodal}.

\section{Approach} \label{sec:approach}

\subsection{GA3C-CADRL}
Recall the RL training process seeks to find the optimal policy, $\pi: \left( \mathbf{s}_t, \, \tilde{\mathbf{s}_t}^o \right) \mapsto \mathbf{u}_t$, which maps from an agent's observation of the environment to a probability distribution across actions.
We use a local coordinate frame as in~\cite{Chen17_IROS,chen_decentralized_2017}, and separate the state of the world in two pieces: information about the agent itself, and everything else in the world.
Information about the agent can be represented in a small, fixed number of variables.
The world, on the other hand, can be full of any number of other objects, or even completely empty.
Specifically, there is one $\mathbf{s}$ vector about the agent itself, and one $\tilde{\mathbf{s}}^o$ vector per other agent in the vicinity:
\begin{align}
	\mathbf{s} & = [d_g, \; v_{pref}, \; \psi, \; r] \label{eqn:agent_state} \\  
	\tilde{\mathbf{s}}^o & = [\tilde{p}_x, \; \tilde{p}_y, \; \tilde{v}_x, \; \tilde{v}_y,
		 \; \tilde{r}, \; \tilde{d}_a, \; \tilde{r}+r] \;  \label{eqn:other_state}, 
\end{align}
where $d_g=||\mathbf{p}_g - \mathbf{p}||_2$ is the agent's distance to goal, and $\tilde{d}_a=||\mathbf{p} - \tilde{\mathbf{p}}||_2$ is the distance to the other agent.

The agent's action space is composed of a speed and change in heading angle.
It is discretized into 11 actions: with a speed of $v_{pref}$ there are 6 headings evenly spaced between $\pm\pi/6$, and for speeds of $\frac{1}{2}v_{pref}$ and 0 the heading choices are $[-\pi/6, 0, \pi/6]$. 
These actions are chosen to mimic real turning constraints of robotic vehicles.

The sparse reward function is defined as
\begin{align} \hspace*{-.1in}
  R_{col}(\mathbf{s}^{jn}) = 
  \begin{cases}
    1 & \text{if $\mathbf{p} = \mathbf{p}_g$} \label{eqn:reward} \\
    -0.25 & \text{if $d_{min} < 0$} \\
    -0.1 + 0.05\cdot d_{min} & \text{if $0 < d_{min} < 0.2$} \\
    0 & \text{otherwise}
  \end{cases}
\end{align}
where $d_{min}$ is the distance to the closest other agent.
Note that we use discount $\gamma < 1$ to encourage efficiency instead of a step penalty.

This RL problem formulation is solved with GA3C in a process we call GA3C-CADRL (GPU/CPU Asynchronous Advantage Actor-Critic for Collision Avoidance with Deep RL).
As opposed to many RL problems that involve a single agent exploring in an environment, the collision avoidance domain often has several agents using the learned policy in each training episode.
Since experience generation is one of the time-intensive parts of training, this work extends GA3C to learn from multiple agents' experiences each episode.
Training batches are filled with a mix of agents' experiences ($\{\mathbf{s}^{jn}_t, \mathbf{u}_t, r_t\}$ tuples) to encourage policy gradients that improve the joint expected reward of all agents.
The extended implementation accounts for agents reaching their goals at different times, and ignores experiences of agents running other policies (e.g. non-cooperative agents).

\subsection{Handling a Variable Number of Agents}

\begin{figure}[t]
	\centering
	\includegraphics [trim=20 100 50 50, clip, width=0.5 \textwidth, angle = 0, page=8]{figures/network_architecture_v4.pdf}
	\caption{LSTM unrolled to show each input. At each decision step, the agent feeds one observable state vector, $\tilde{\mathbf{s}}^o_i$, for each nearby agent, into a LSTM cell sequentially. LSTM cells store the pertinent information in the hidden states, $h_i$. The final hidden state, $h_n$, encodes the entire state of the other agents in a fixed-length vector, and is then fed to the feedforward portion of the network. The order of agents is sorted by decreasing distance to the ego agent, so that the closest agent has the most recent effect on $h_n$.}
	\label{fig:lstm}
%
	\centering
	\includegraphics [trim=40 40 50 50, clip, width=0.5 \textwidth, angle = 0, page=1]{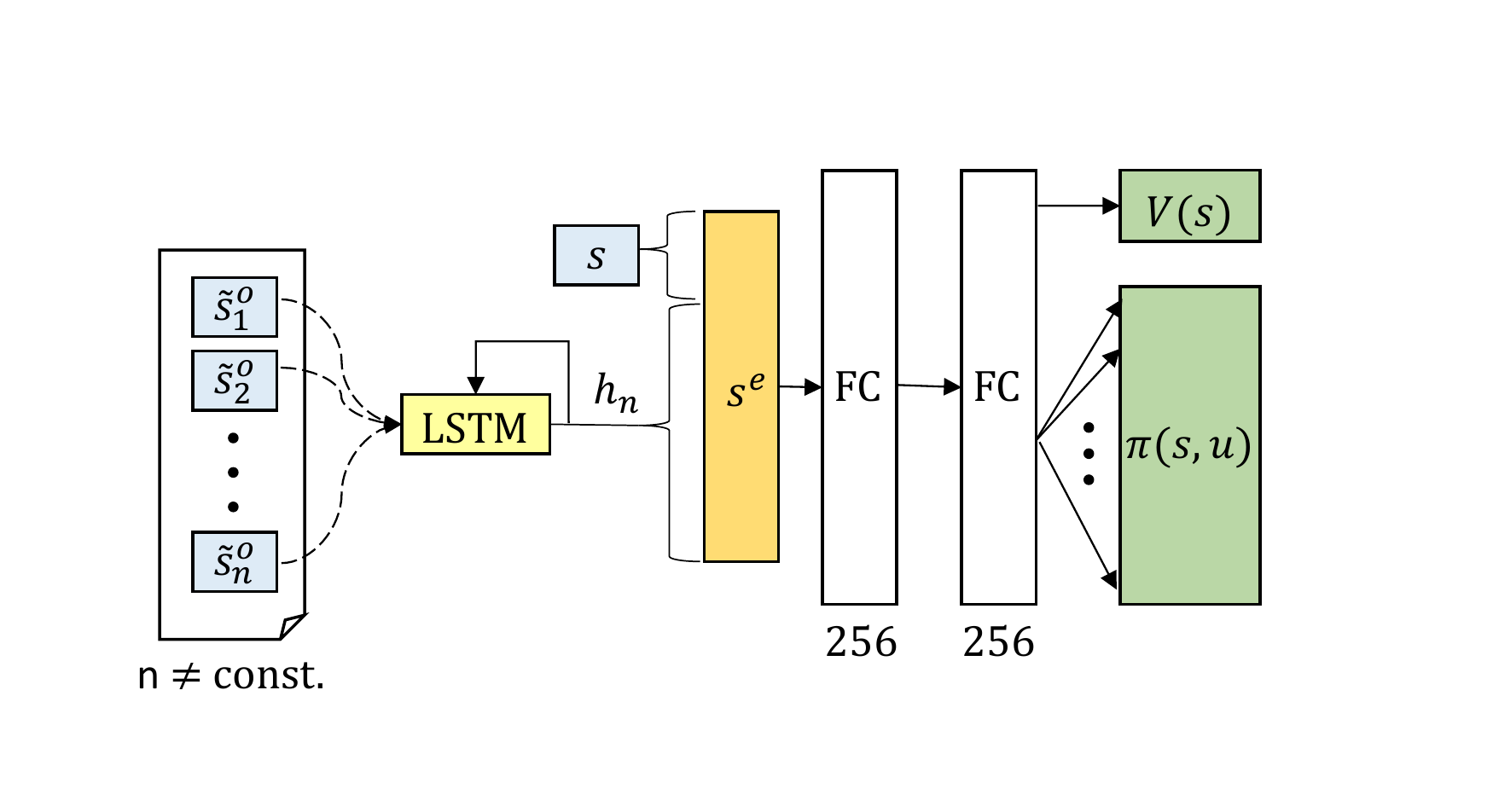}
	\caption{Network Architecture. Observable states of nearby agents, $\tilde{\mathbf{s}}^o_i$, are fed sequentially into the LSTM, as unrolled in~\cref{fig:lstm}. The final hidden state is concatenated with the agent's own state, $\mathbf{s}$, to form the golden vector, $\mathbf{s}^e$. For any number of agents, $\mathbf{s}^e$ contains the agent's knowledge of its own state and the state of the environment. The encoded state is fed into two fully-connected layers (FC). The outputs are a scalar value function (green, top) and policy represented as a discrete probability distribution over actions (green, bottom).}
	\label{fig:nn_arch}
\end{figure}

Recall that one key limitation of many learning-based collision avoidance methods is that the feedforward NNs typically used require a fixed-size input.
Convolutional and max-pooling layers are useful for feature extraction and can modify the input size, but still convert a fixed-size input into a fixed-size output.
Recurrent NNs, where the output is produced from a combination of a stored cell state and an input, accept an arbitrary-length sequence to produce a fixed-size output.
Long short-term memory (LSTM)~\cite{hochreiter1997long} is recurrent architecture with advantageous properties for training\footnote{In practice, TensorFlow's LSTM implementation requires a known maximum sequence length, but this can be set to something bigger than the number of agents agents ever expected (e.g. 20)}.

Although LSTMs are often applied to time sequences of data (e.g. pedestrian motion prediction~\cite{alahi2016social}), this paper leverages their ability to encode a sequence of information that is not time-dependent.
In this work, we treat the variable number of $\tilde{\mathbf{s}}^o$ vectors as a sequence of inputs that encompass everything the agent knows about the rest of the world.
At each decision step, the agent feeds each $\tilde{\mathbf{s}}^o$ into a LSTM cell sequentially, as in~\cref{fig:lstm}.
That is, the LSTM initially has an empty state and accepts $\tilde{\mathbf{s}}^o_1$ to generate $h_1$, then feeds $h_1$ and $\tilde{\mathbf{s}}^o_2$ to produce $h_2$, and so on.
As agents' states are fed in, the LSTM stores the pertinent information in its hidden state, and forgets the less important parts of the input.
After inputting the final agent's state, we can interpret the LSTM's final hidden state as a fixed-length, encoded state of the world, for that decision step.

Given a sufficiently large hidden state vector, there is enough space to encode a large number of agents' states without the LSTM having to forget anything relevant.
In the case of a large number of agent states, to mitigate the impact of the agent forgetting the early states, the states are fed in reverse order of distance to the agent, meaning the closest agents (fed last) should have the biggest effect on the final hidden state, $h_n$.

The idea is visualized in~\cref{fig:nn_arch}, where the blue box labeled is the agent's own state, $\mathbf{s}$, and the group of blue boxes is the $n$ other agents' observable states, $\tilde{\mathbf{s}}^o_i$.
After passing the $n$ other agents' observable states into the LSTM, the agent's own state is concatenated with $h_n$, and this new vector becomes the input to a typical feedforward DNN with 2 fully-connected layers.
The network produces two output types: a scalar state value, and policy composed of a probability for each action in the discrete action space.
During the backpropagation training process, the LSTM's weights are updated to learn how to represent the variable number of other agents in a fixed-length $h$ vector.

\subsection{Training the Policy}
The original CADRL and SA-CADRL (Socially Aware CADRL) algorithms used several clever tricks to enable convergence when training the networks.
Specifically, forward propagation of other agent states for $\Delta t$ seconds was a critical component that required tuning, but does not represent agents' true behaviors.
Other details include separating experiences into successful/unsuccessful sets to focus the training on cases where the agent could improve.
The new GA3C-CADRL formulation is more general, and does not require such assumptions or modifications.

In this work, to train the model, the network weights are first initialized in a supervised learning phase, which converges in $<5$ minutes.
The initial training is done on a large set of state-action-value pairs from an existing CADRL solution, where the network loss combines square-error loss on the value output and cross-entropy loss on the policy output.
This training set is released to the public to aid in network initialization.

The initialization step is necessary to enable any possibility of later generating useful RL experiences (non-initialized agents wander randomly and probabilistically almost never obtain positive reward). 
Agents running the initialized GA3C-CADRL policy reach their goals reliably when there are no interactions with other agents.
However, the policy after this supervised learning process still performs poorly in collision avoidance.
This observation contrasts with CADRL, in which the initialization step was sufficient to learn a policy that performs comparably to existing reaction-based methods, due to relatively-low dimension value function combined with manual propagation of states.
Key reasons behind this contrast are the reduced structure in the GA3C-CADRL formulation (no forward propagation), and that the algorithm is now learning both a policy and value function (as opposed to just a value function), since the policy has an order of magnitude higher dimensionality than a scalar value function.

To improve the solution with RL, experiences are generated from simulations of randomly-generated scenarios.
These scenarios include several agents trying to get to their randomly-positioned goals, running a random assortment of policies (Non-Cooperative, Zero Velocity, or the learned GA3C-CADRL policy at that iteration), but only experiences from agents using the GA3C-CADRL policy are fed back to the trainer.
Agent parameters vary between $r~\in~[0.2,\, 0.8]$m, and $v_{pref} \in [0.5, \, 2.0]$m/s, chosen to be near pedestrian values.

An important benefit of the new framework is that the policy can be trained on scenarios involving any number of agents, whereas the maximum number of agents had to be defined ahead of time with CADRL/SA-CADRL\footnote{Experiments suggest this number should be below about 6 for convergence}.
This work begins the RL phase with 2-4 agents in the environment, so that the policy learns the idea of collision avoidance in reasonably simple domains.
Upon convergence, a second RL phase begins with 2-10 agents in the environment.
\section{Results} \label{sec:results}
\subsection{Computational Details}

The DNNs in this work were implemented with TensorFlow~\cite{abadi2016tensorflow} in Python.
A query of the new GA3C-CADRL network only requires the current state vector, while the previous approach queried the value of a batch of future propogated states to choose which action is best. 
Accordingly, this implementation is much more efficient than the networks in~\cite{chen_decentralized_2017,Chen17_IROS}: a single query takes on average 0.4-0.5ms on a i7-7700K CPU, $\sim$20x faster than before.
Note that a GPU is not required for fast execution of a trained model.

In total, the RL converges in about 12 hours (after $2~\cdot~10^{6}$ episodes) for the multi-agent, LSTM network on a computer with an NVIDIA GTX1060 graphics card.
A limiting factor of the training time is the low learning rate required for stable training.
Recall that the previous approach took 8 hours to train a 4-agent value network, but now in a similar amount of time (albeit using many more episodes) the network learns both the policy and value function, and without being provided any structure about the other agents behaviors.
The increased number of episodes required can also be attributed to the stark contrast in initial policies upon starting RL: CADRL was fine-tuning a decent policy, whereas GA3C-CADRL learns collision avoidance entirely in the RL phase.

After initialization, the agents receive on average 0.15 reward per episode.
After RL phase 1 (converges in $1.5\cdot10^6$ episodes), they average 0.90 reward per episode. 
When RL phase 2 begins, the average reward drops to 0.85 initially since the domain becomes much harder ($n_{max}$ increases from 4 to 10), and then increases until converging at 0.93 (after a total of $1.9\cdot10^6$ episodes).
Reward is computed as the sum of the rewards accumulated in each episode, averaged across all GA3C-CADRL agents in that episode.
Reward is just a measure of success/failure, as it does not include the discount factor and thus is not indicative of time efficiency.
Because the maximum receivable reward on any episode is 1, an average reward $<1$ implies there are some collisions (or other penalized behavior) even after convergence.
This is expected, as agents sample from their policy distributions when selecting actions in training, so there is always a non-zero probability of choosing a sub-optimal action in training.
Later, when executing a trained policy, agents select the action with highest probability.

Key parameter values include: learning rate $L_r = 2\cdot 10^{-5}$, entropy coefficient $\beta = 1 \cdot 10^{-4}$, discount $\gamma = 0.97$, training batch size $b_s = 100$, and we use the Adam optimizer~\cite{kingma2014adam}.

\subsection{Simulation Results}

\begin{figure*}[t]
	\centering
	\begin{subfigure}{0.99\textwidth}
		\centering
		\includegraphics [trim=20 0 50 60, clip, width=0.24 \textwidth, angle = 0]{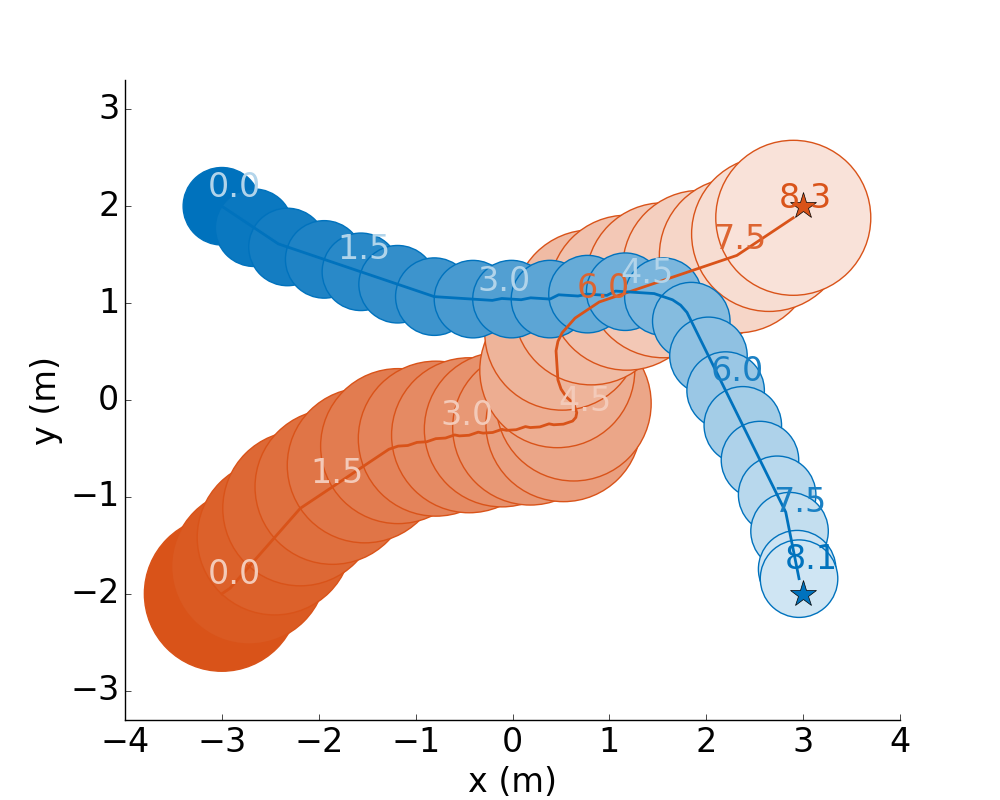}
		\includegraphics [trim=20 0 50 60, clip, width=0.24 \textwidth, angle = 0]{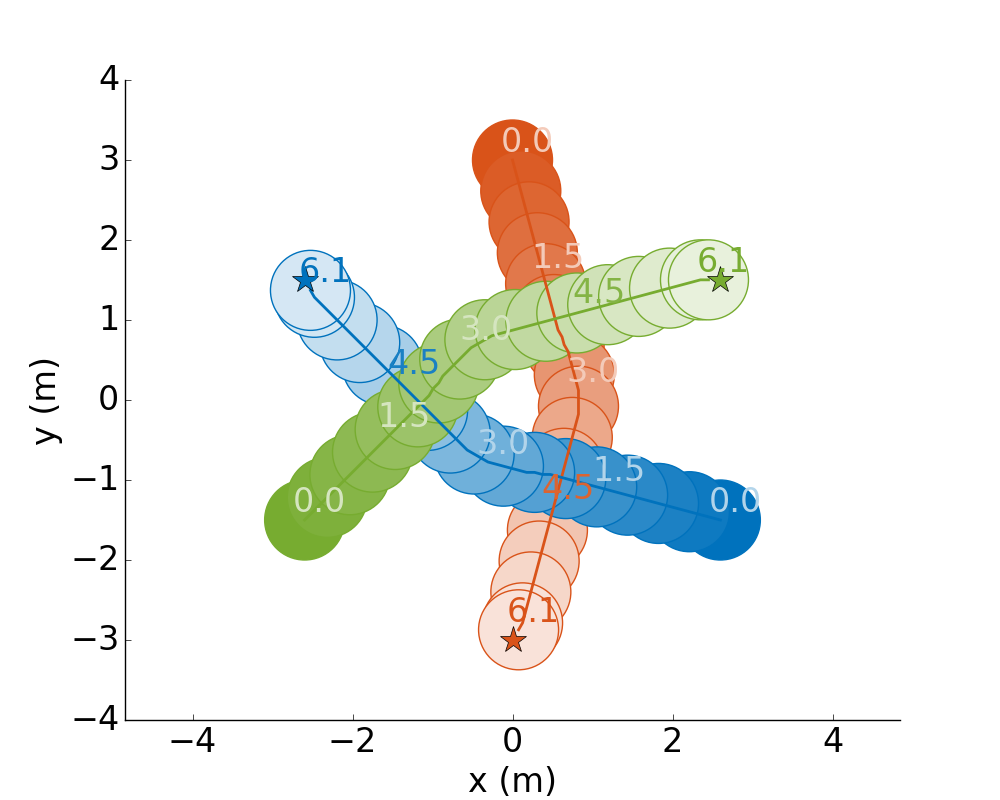}
		\includegraphics [trim=20 0 50 60, clip, width=0.24 \textwidth, angle = 0]{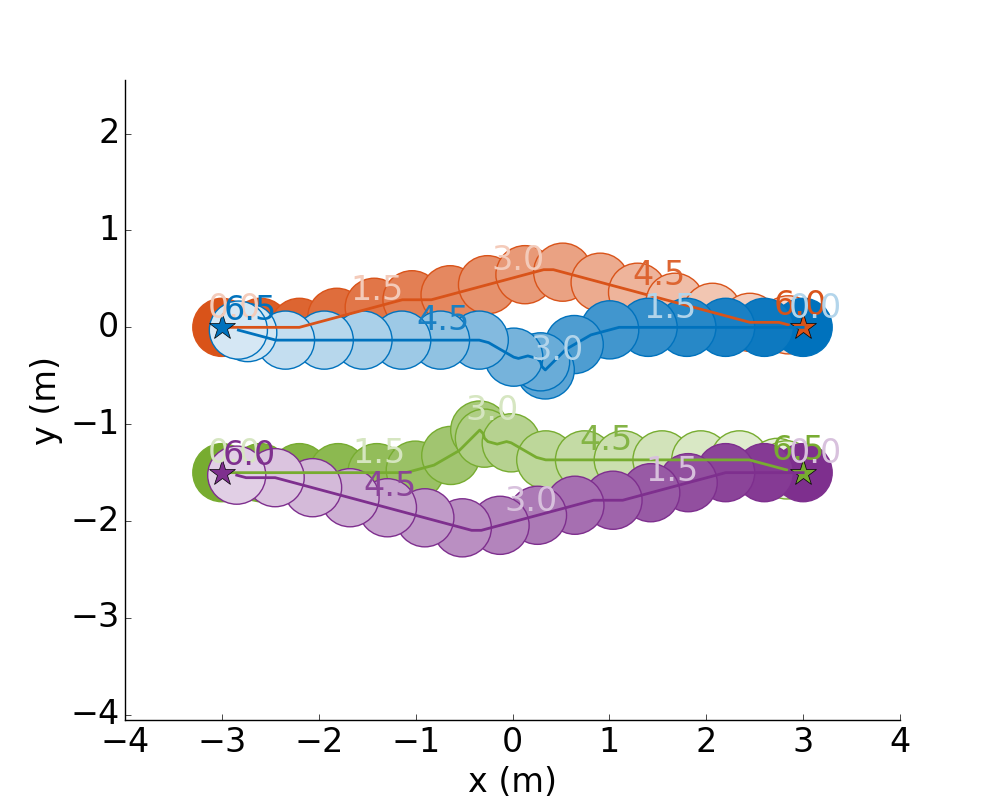}
		\includegraphics [trim=20 0 50 60, clip, width=0.24 \textwidth, angle = 0]{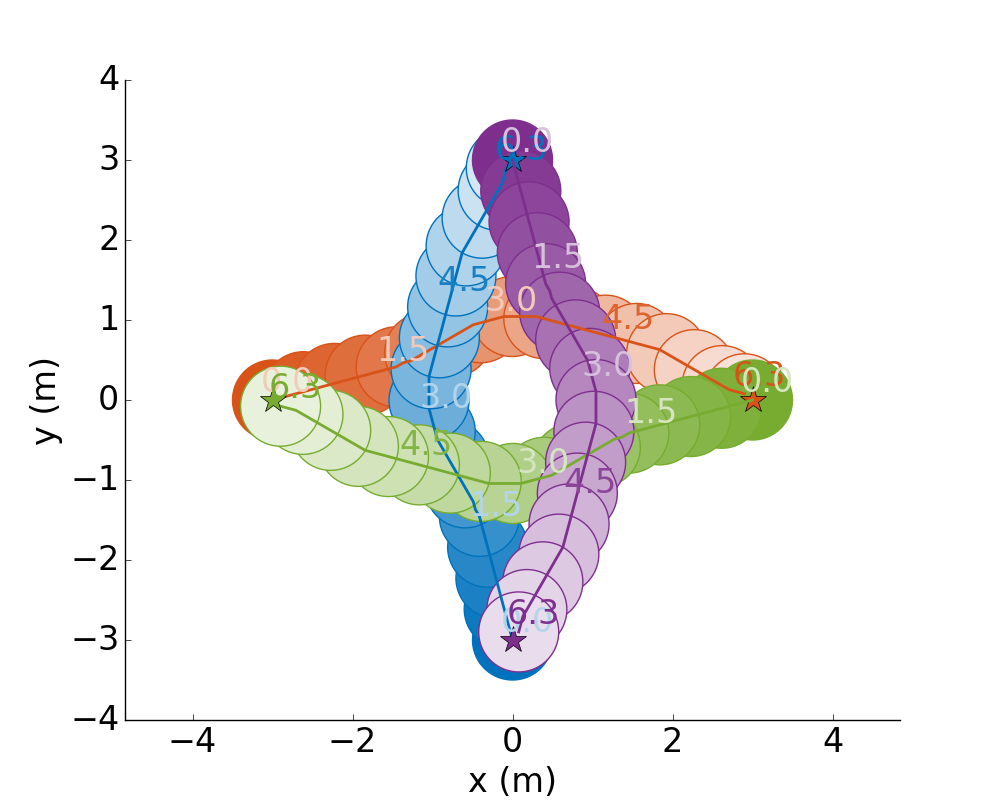}
		\caption{GA3C-CADRL trajectories with $n\in[2,3,4]$ agents}
		\label{fig:ga3c_cadrl_traj_2_agent} 
	\end{subfigure}
	\begin{subfigure}{0.99\textwidth}
		\centering
		\includegraphics [trim=20 0 50 60, clip, width=0.24 \textwidth, angle = 0]{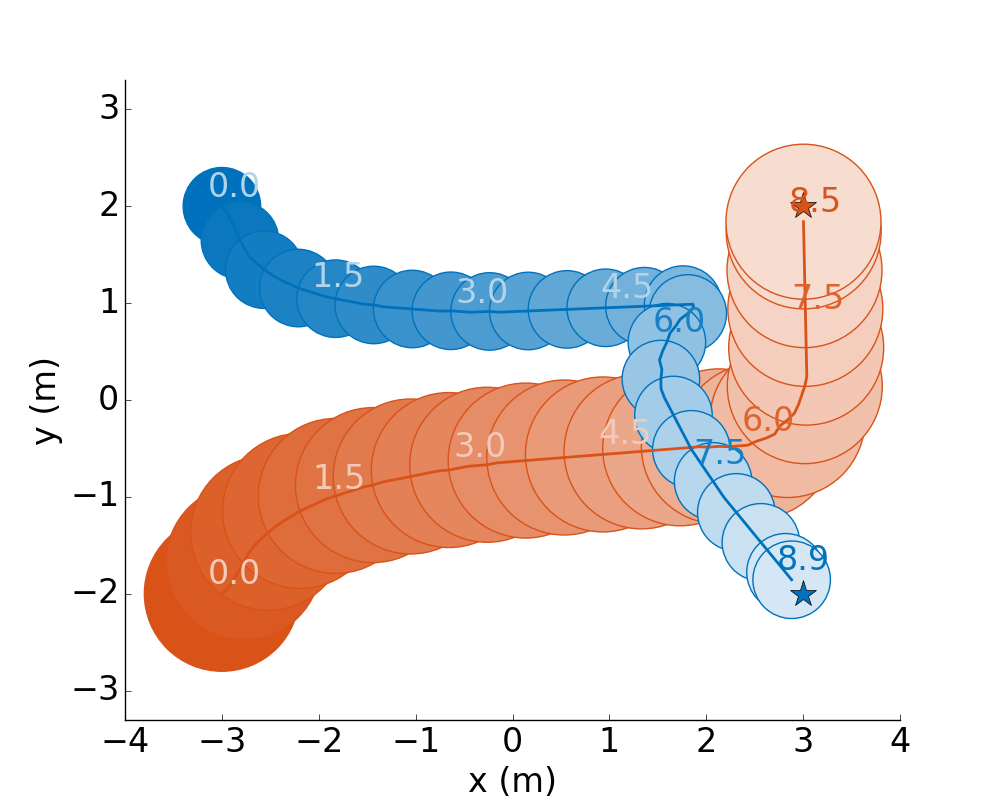}
		\includegraphics [trim=20 0 50 60, clip, width=0.24 \textwidth, angle = 0]{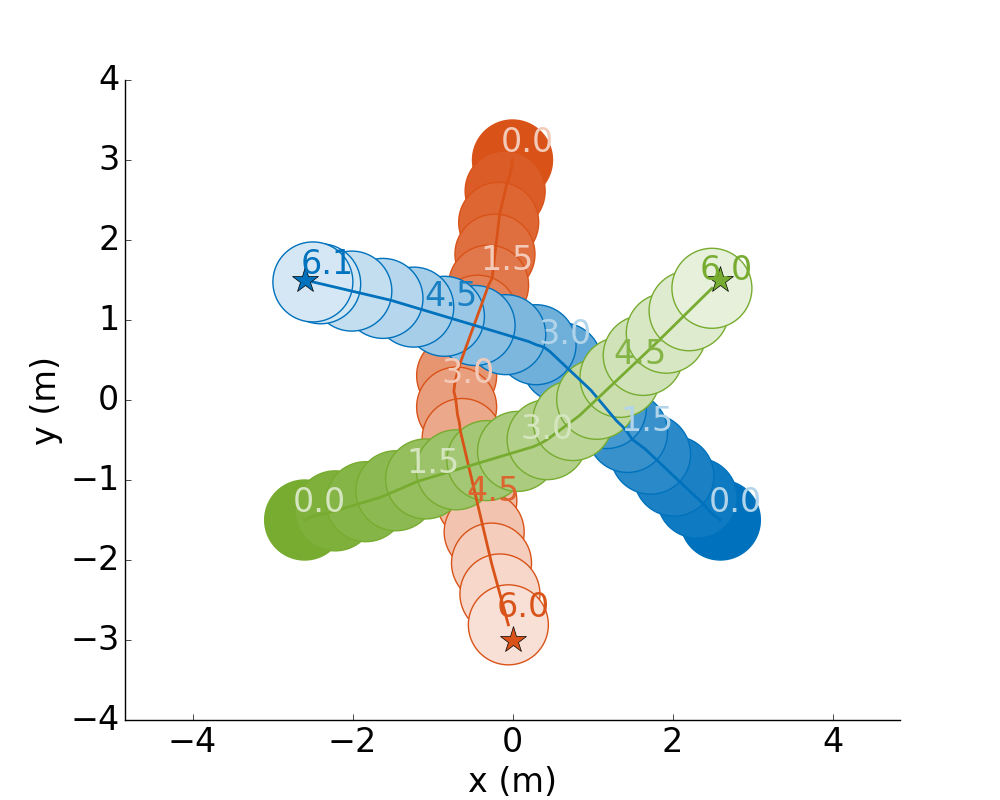}
		\includegraphics [trim=20 0 50 60, clip, width=0.24 \textwidth, angle = 0]{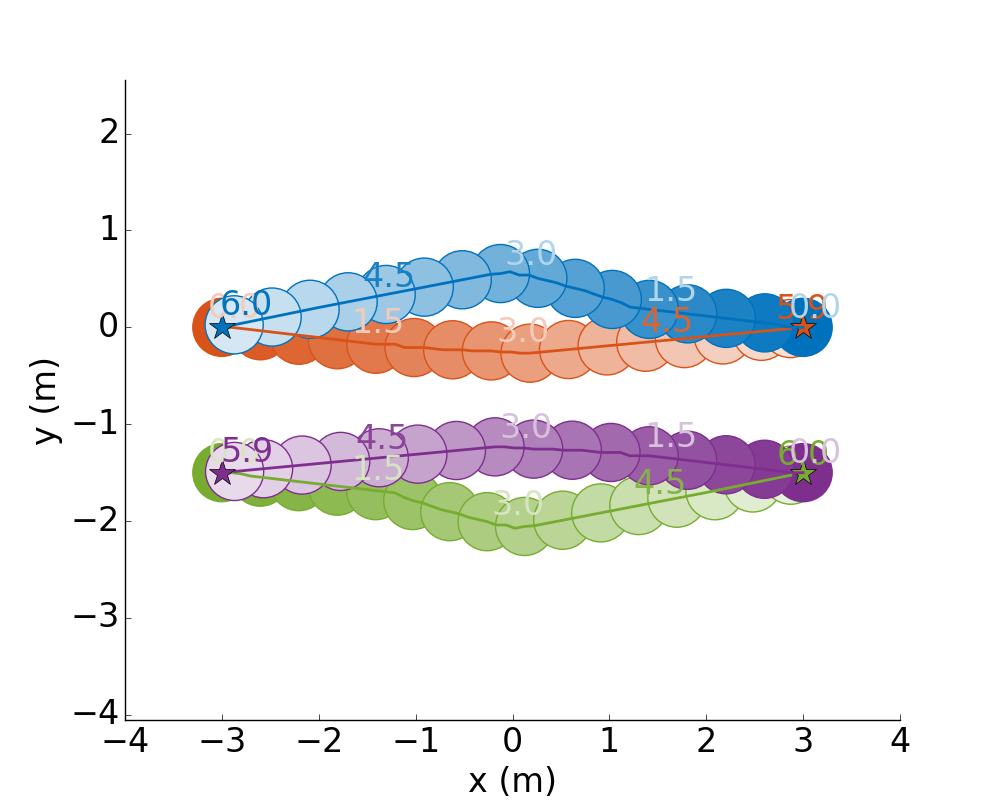}
		\includegraphics [trim=20 0 50 60, clip, width=0.24 \textwidth, angle = 0]{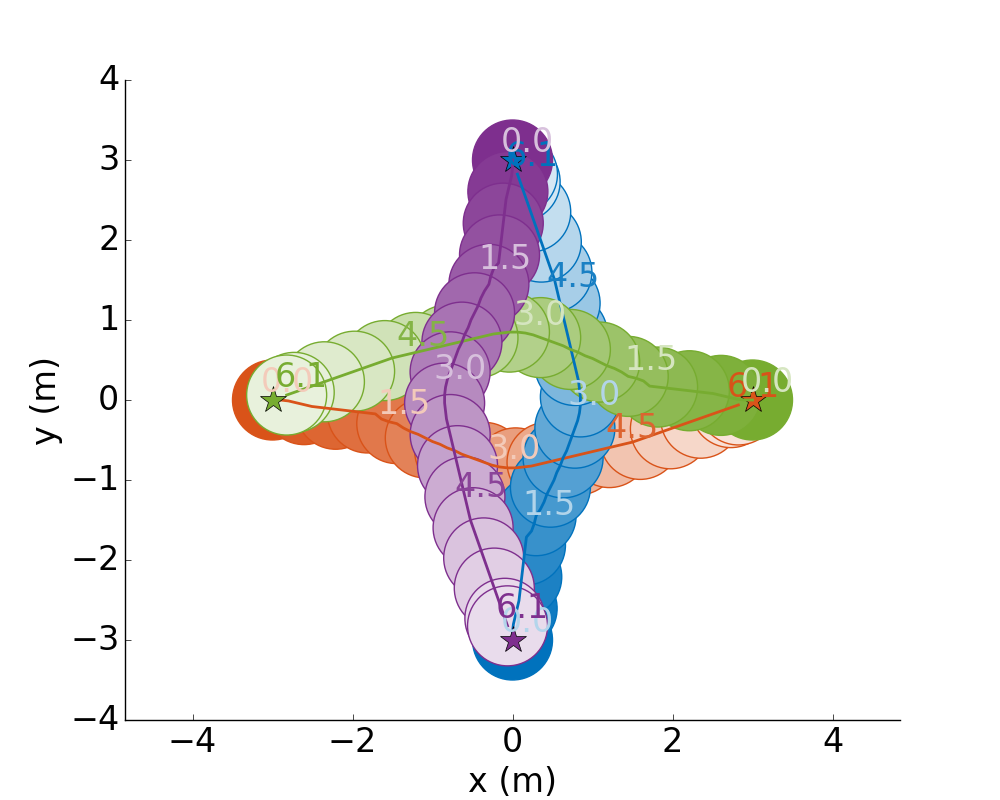}
		\caption{SA-CADRL trajectories with $n\in[2,3,4]$ agents}
		\label{fig:cadrl_traj_2_agent} 
	\end{subfigure}
	\caption{Scenarios with $n\leq4$ agents. The top row shows agents executing GA3C-CADRL-10, and the bottom row shows same scenarios with agents using SA-CADRL. Circles lighten as time increases, and the numbers represent the time at agent's position. GA3C-CADRL agents are slightly less efficient, as they reach their goals slightly slower than SA-CADRL agents. However, the overall behavior is similar, and the more general GA3C-CADRL framework generates desirable behavior without many of the assumptions from SA-CADRL.}
	\label{fig:234agent_traj}
	\vskip-0.1in
\end{figure*}

Although the original 2-agent CADRL algorithm~\cite{chen_decentralized_2017} was also shown to scale to multi-agent scenarios, its minimax implementation is limited in that it only considers one neighbor at a time as described in~\cite{Chen17_IROS}.
For that reason, this work focuses the comparison against SA-CADRL which has better multi-agent properties - the policy used for comparison is the same one that was used on the robotic hardware in~\cite{Chen17_IROS}.
That particular policy was trained with some noise in the environment ($\mathbf{p} = \mathbf{p}_{actual} + \sigma$) which led to slightly poorer performance than the ideally-trained network as reported in the results of~\cite{Chen17_IROS}, but more acceptable hardware performance.

The version of the new GA3C-CADRL policy after RL phase 2 is denoted GA3C-CADRL-10, as it was trained in scenarios of up to 10 agents.
To create a more fair comparison with SA-CADRL which was only trained with up to 4 agents, let GA3C-CADRL-4 denote the policy after RL phase 1 (which only involves scenarios of up to 4 agents). 
Recall GA3C-CADRL-4 can still be naturally implemented on $n>4$ agent cases, whereas SA-CADRL can only accept up to 3 nearby agents' states regardless of $n$.

\subsubsection{$n \leq 4$ agents}
The previous approach (SA-CADRL) is known to perform well on scenarios involving a few agents ($n \leq 4$), as its trained network can accept up to 3 other agents' states as input.
Therefore, the goal is to confirm that the new algorithm can still perform comparably.
This is not a trivial check, as the new algorithm is not provided with any structure/prior about the world's dynamics, so the learning is more difficult.

Trajectories are visualized in~\cref{fig:234agent_traj}: the top row shows scenarios with agents running the new policy (GA3C-CADRL-10), and the bottom row shows agents in identical scenarios but using the old policy (SA-CADRL).
The colors of the circles (agents) lighten as time increases and the circle size represents agent radius (not constant).
The trajectories generally look similar for both algorithms, with SA-CADRL being slightly more efficient.
A rough way to assess efficiency in these plotted paths is time indicated when the agents reach their goals.

Although it is easy to pick out interesting pros/cons for any particular scenario, it is more useful to draw conclusions after aggregating over a large number of randomly-generated cases.
Thus, we created test sets of 500 random scenarios, defined by ($p_{start}$, $p_{goal}$, $r$, $v_{pref}$) per agent, for many different numbers of agents.
Each algorithm is evaluated on the same 500 test cases.
The comparison metrics are the percent of cases with a collision, percent of cases where an agent gets stuck and doesn't reach the goal, and of the remaining cases where all algorithms were successful, the average extra time to goal, $\bar{t}^{e}_g$ beyond a straight path at $v_{pref}$\footnote{This evaluation could be slightly unfair to the algorithm that has fewer failures, because it ignores potentially highly efficient cases for one algorithm that led to failures by another algorithm}.
These metrics provide measures of efficiency and safety.

Aggregated results in~\cref{tab:multi_agent_stats} suggest that both of the new GA3C-CADRL policies perform comparably to, though slightly worse than, SA-CADRL with $n \leq 4$ agents in the environment.
SA-CADRL has the lowest $\bar{t}^{e}_g$, and the agents rarely fail in these relatively simple scenarios.
The difference between GA3C-CADRL-4 and GA3C-CADRL-10 is small for $n \leq 4$, which makes sense because GA3C-CADRL-4 converged after being trained in scenarios involving few agents.
The minor improvement could be explained by GA3C-CADRL-10's LSTM weights, which would have seen more examples of various numbers of agents, and therefore are better trained to represent $n$ observations in a fixed-size $h$ vector.

\begin{figure}[t]
	\vskip-0.1in
	\centering
	\includegraphics [trim=0 60 0 0, clip, angle=0, width=0.8\columnwidth, keepaspectratio]{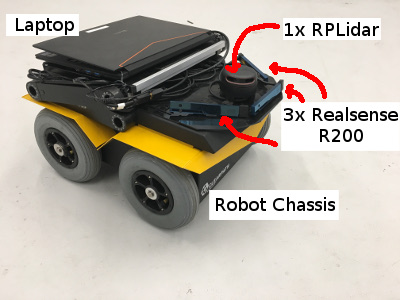}
	\caption{Robot hardware. The compact, low-cost ($<~\$1000$) sensing package uses a single 2D Lidar and 3 Intel Realsense R200 cameras. The total sensor and computation assembly is less than 3 inches tall, leaving room for cargo.}
	\label{fig:robot_sensors} 
	\vskip-0.25in
\end{figure}

\subsubsection{$n > 4$ agents}
A real robot will likely encounter more than 3 pedestrians at a time in a busy environment.
Recall SA-CADRL cannot accept more than 3 other agents' states as input, so the approach taken here is to supply only the closest 3 agents' states in crowded scenarios.
The number of agents is not limited in GA3C-CADRL, as any number of agents can be fed into the LSTM and the final hidden state can still be taken as a representation of the entire world configuration.

\begin{figure*}[t]
	\centering
	\begin{subfigure}{0.32\textwidth}
		\centering
		\includegraphics [trim=20 0 50 40, clip, width=\textwidth, angle = 0]{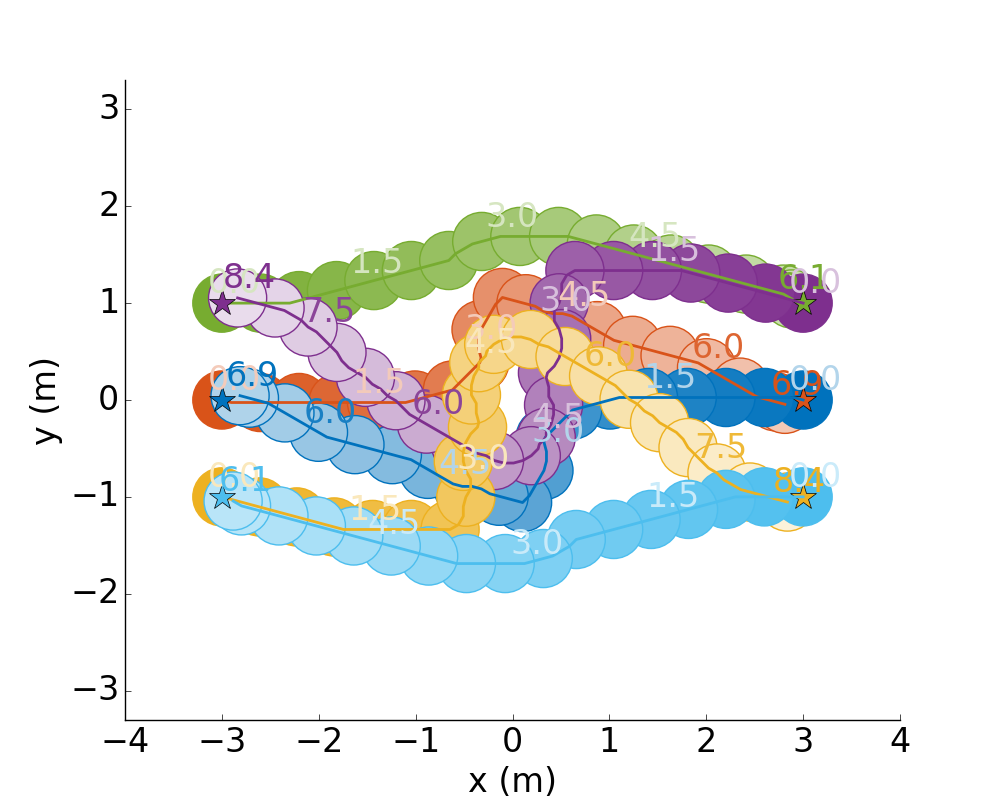}
		\caption{GA3C-CADRL: 3 pair swaps}
		\label{fig:ga3c_cadrl_traj_6_agent_swap} 
	\end{subfigure}
	\begin{subfigure}{0.32\textwidth}
		\centering
		\includegraphics [trim=20 0 50 40, clip, width=\textwidth, angle = 0]{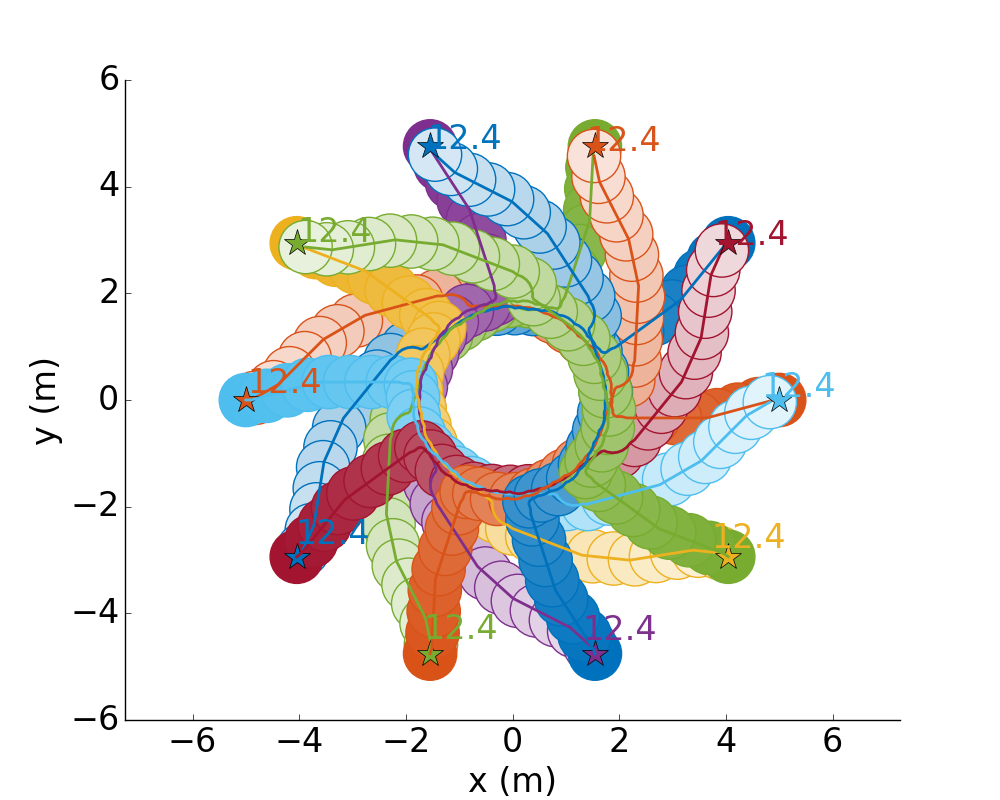}
		\caption{GA3C-CADRL: 10-agent circle}
		\label{fig:ga3c_cadrl_traj_10_agent_circle} 
	\end{subfigure}
	\begin{subfigure}{0.32\textwidth}
		\centering
		\includegraphics [trim=20 0 50 40, clip, width=\textwidth, angle = 0]{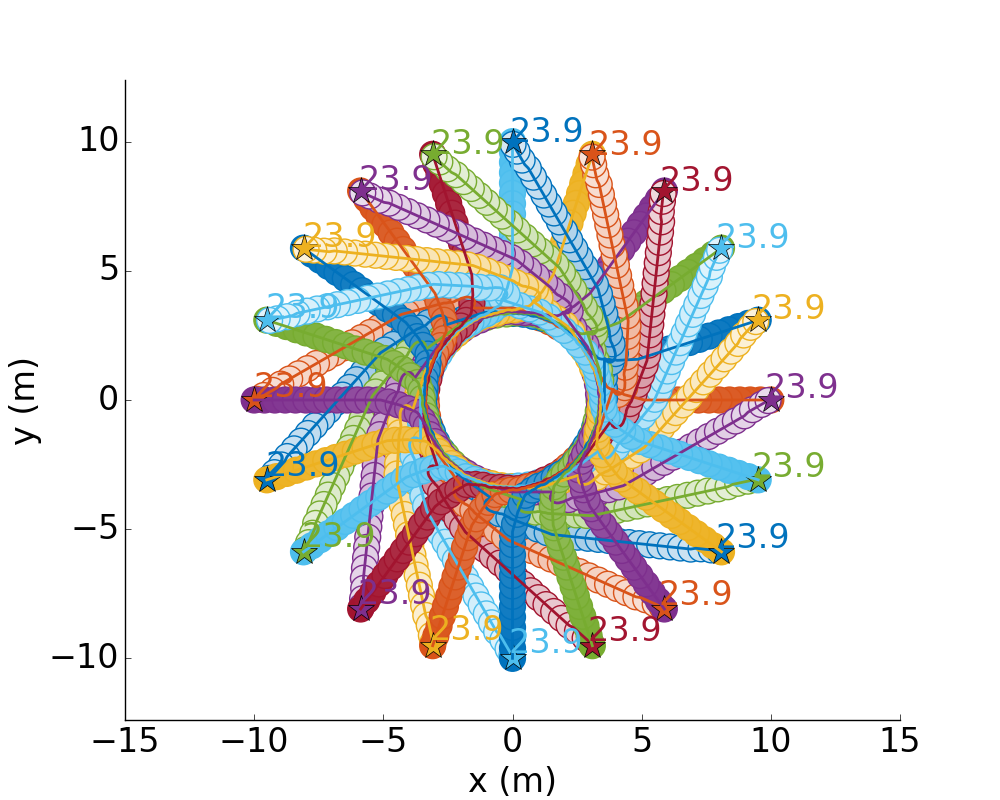}
		\caption{GA3C-CADRL: 20-agent circle}
		\label{fig:ga3c_cadrl_traj_20_agent_circle} 
	\end{subfigure}
	\begin{subfigure}{0.32\textwidth}
		\centering
		\includegraphics [trim=20 0 50 40, clip, width=\textwidth, angle = 0]{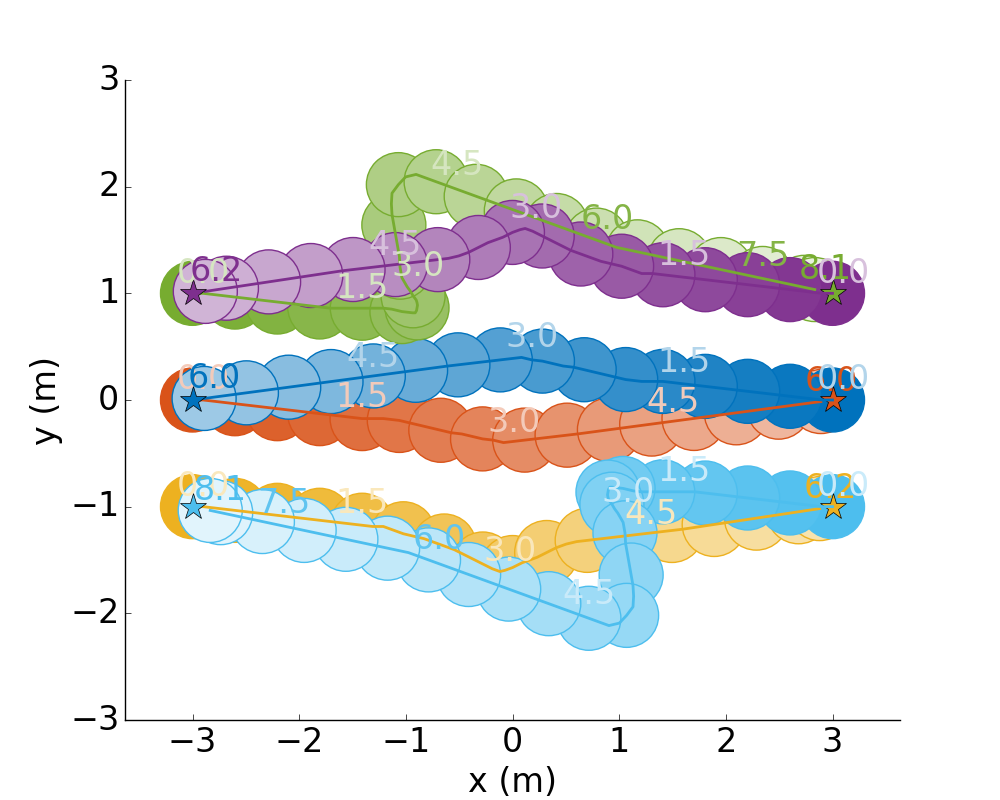}
		\caption{SA-CADRL: 3 pair swaps}
		\label{fig:sa_cadrl_traj_6_agent_swap} 
	\end{subfigure}
	\begin{subfigure}{0.32\textwidth}
		\centering
		\includegraphics [trim=20 0 50 40, clip, width=\textwidth, angle = 0]{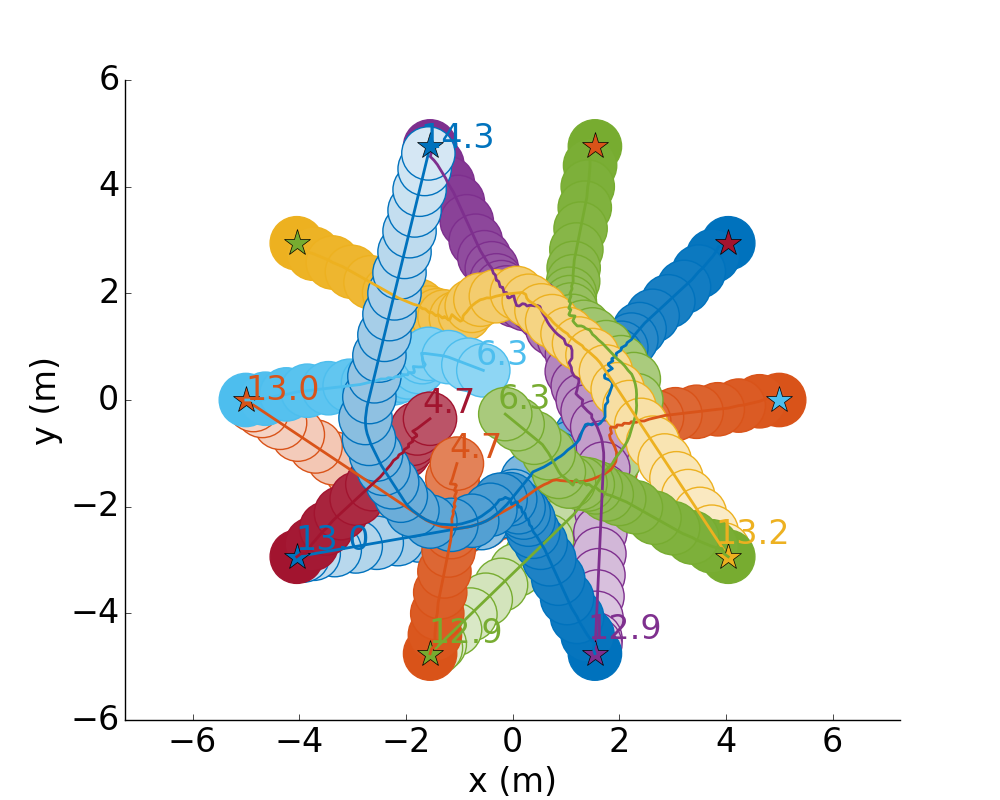}
		\caption{SA-CADRL: 10-agent circle}
		\label{fig:sa_cadrl_traj_10_agent_circle} 
	\end{subfigure}
	\begin{subfigure}{0.32\textwidth}
		\centering
		\includegraphics [trim=20 0 50 40, clip, width=\textwidth, angle = 0]{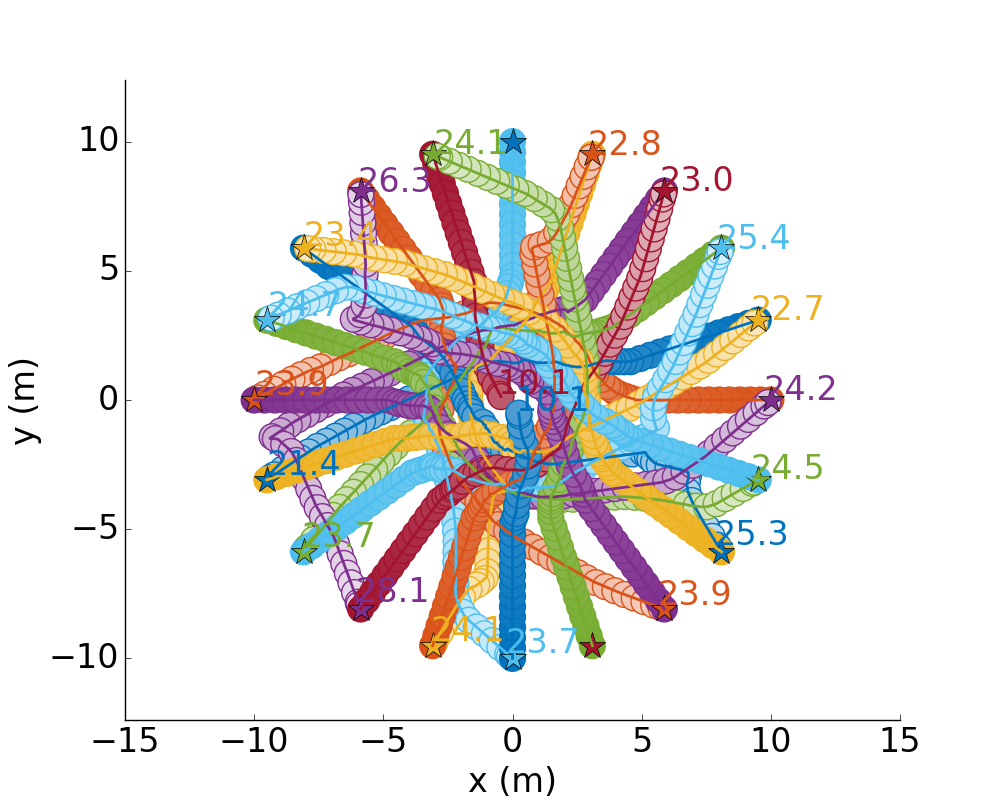}
		\caption{SA-CADRL: 20-agent circle}
		\label{fig:sa_cadrl_traj_20_agent_cirlce} 
	\end{subfigure}
	\caption{Scenarios with $n>4$ agents.
	In the 6-agent swap~\cref{fig:sa_cadrl_traj_6_agent_swap,fig:ga3c_cadrl_traj_6_agent_swap}, GA3C-CADRL agents exhibit interesting multi-agent behavior: the orange and yellow agents form a pair while passing the blue and purple agents.
	SA-CADRL agents reach the goal more quickly than GA3C-CADRL agents, but such multi-agent behavior is a result of GA3C-CADRL agents having the capacity to observe all of the other 5 agents each time step.
	In other scenarios, GA3C-CADRL agents successfully navigate the 10- and 20-agent circles, whereas some SA-CADRL agents collide (i.e. orange/red and blue/green in~\cref{fig:sa_cadrl_traj_10_agent_circle}, blue/red near (0,0) in~\cref{fig:sa_cadrl_traj_20_agent_cirlce}).}
	\label{fig:more_than_4agent_traj}
	\vskip-0.2in
\end{figure*}

Even in $n>4$-agent environments, interactions still often only involve a couple of agents at a time.
Some specific cases where there truly are many-agent interactions are visualized in~\cref{fig:more_than_4agent_traj}.
In the 6-agent swap (left), GA3C-CADRL agents exhibit interesting multi-agent behavior: the orange and yellow agents form a pair while passing the blue and purple agents.
This phenomenon leads to a particularly long path for yellow and purple, but also allows the outside agents, green and light blue, to not deviate as much from a straight line.
In contrast, in SA-CADRL the green agent starts moving right and downward, until dark blue becomes one of the closest 3 neighbors.
Green then makes an escape maneuver and passes purple on the outside.
In this case, SA-CADRL agents reach the goal more quickly than GA3C-CADRL agents, but the interesting multi-agent behavior is a result of GA3C-CADRL agents having the capacity to observe all of the other 5 agents each time step, rather than SA-CADRL which just uses the nearest 3 neighbors.

GA3C-CADRL agents successfully navigate the 10- and 20-agent circles (antipodal swaps), whereas several SA-CADRL agents get stuck or collide\footnote{Note there is not perfect symmetry in these SA-CADRL cases: small numerical fluctuations affect the choice of the closest agents, leading to slightly different actions for each agent. And after a collision occurs with a pair of agents, symmetry will certainly be broken for future time steps.}

Statistics across 500 random cases of 5,6,8, and 10 agents are listed in~\cref{tab:multi_agent_stats}.
The performance gain by using GA3C-CADRL becomes stronger as the number of agents in the environment increases.
The performance of each algorithm is similar when $n = 5$, but a large change occurs at $n=6$, with a 5x reduction in failed cases and a shorter $\bar{t}^{e}_g$ for GA3C-CADRL-10 agents over SA-CADRL.
GA3C-CADRL-10 outperforms the other algorithms when $n=8$ and $n=10$ as well.
GA3C-CADRL-10's percent of success remains above 95\% across any $n<10$, whereas SA-CADRL drops to under 80\%.
It is worth noting that SA-CADRL agents' failures are more often a result of getting stuck rather than colliding with others, however neither outcomes are desirable.
The domain size of $n=10$ agent scenarios is set to be larger (6x6 vs. 4x4 m) than cases with smaller $n$ to demonstrate cases where $n$ is large but the world is not necessarily more densely populated with agents.
Accordingly, GA3C-CADRL agents' probability of success is actually slightly better with $n=10$ vs. $n=8$, even though there are more agents.

The results comparing just the two GA3C-CADRL policies demonstrate the benefit of the second RL training phase, as there is a large decrease in failed cases and a slight decrease in $\bar{t}^{e}_g$ after training and converging in every one of the $n \leq 10$ environments.
The ability for GA3C-CADRL to retrain in complex scenarios after convergence in simple scenarios, and yield a significant performance increase, is a key benefit of the new framework.
This result suggests there could be other types of complexities in the environment (beyond increasing $n$) that the general GA3C-CADL framework could also learn about after being initially trained on simple scenarios.

\begin{table*}[t]
	\centering
	\caption[]{Performance of SA-CADRL (old) and GA3C-CADRL (new) algorithms on the same 500 random test cases. Average extra time to goal, $\bar{t}^{e}_{g}$, is computed on the test cases where no agents collided or got stuck with either algorithm.
	GA3C-CADRL-10 performs comparably to SA-CADRL for $n\leq4$ and outperforms SA-CADRL significantly for large $n$.}
	\begin{tabular}{|c|c||c|c|c||c|c|c||}
	  \hline
	  \multicolumn{2}{|c||}{Test Case Setup} &
	    \multicolumn{3}{c||}{Extra time to goal $\bar{t}^{e}_{g}$ (s) (Avg / 75th / 90th percentile]} &
	    \multicolumn{3}{c||}{\% failures  (\% collisions / \% stuck)} \\ \hline
	  \# agts. & size (m) & SA-CADRL & GA3C-CADRL-4 & GA3C-CADRL-10 & SA-CADRL & GA3C-CADRL-4 & GA3C-CADRL-10 \\
	  \hline
	  2 & 4 x 4 & \textbf{0.28 / 0.47 / 0.80} 	& 0.28 / 0.46 / 0.87 	& 0.26 / 0.44 / 0.84 			& \textbf{0.0 (0.0 / 0.0)} 	& 0.2 (0.0 / 0.2)	 & \textbf{0.0 (0.0 / 0.0)} \\
	  3 & 4 x 4 & \textbf{0.35 / 0.57 / 0.97} 	& 0.46 / 0.83 / 1.37 	& 0.38 / 0.64 / 1.14		 	& 0.2 (0.0 / 0.2)		 	& 0.4 (0.4 / 0.0)	 & \textbf{0.0 (0.0 / 0.0)} \\
	  4 & 4 x 4 & 0.63 / 0.97 / 1.39 			& 0.68 / 1.14 / 1.69 	& \textbf{0.55 / 0.95 / 1.51} 	& 2.0 (0.0 / 2.0) 			& 1.8 (0.6 / 1.2)	 & \textbf{1.0 (0.8 / 0.2)} \\
	  5 & 4 x 4 & \textbf{0.68 / 1.05 / 1.59} 	& 0.80 / 1.23 / 1.64 	& 0.73 / 1.08 / 1.56			& 2.8 (0.0 / 2.8) 			& 2.4 (1.8 / 0.6)	 & \textbf{1.2 (1.0 / 0.2)} \\
	  6 & 4 x 4 & 0.91 / 1.31 / 1.75 			& 0.98 / 1.38 / 1.86 	& \textbf{0.87 / 1.23 / 1.66} 	& 9.4 (1.6 / 7.8) 			& 3.4 (2.8 / 0.6)	 & \textbf{1.8 (0.8 / 1.0)} \\
	  8 & 4 x 4 & 1.53 / 2.09 / 2.72 			& 1.20 / 1.62 / 2.17 	& \textbf{1.12 / 1.57 / 2.00} 	& 15.2 (1.6 / 13.6) 		& 7.8 (5.8 / 2.0)	 & \textbf{4.2 (3.2 / 1.0)} \\
	  10 & 6 x 6 & 1.33 / 1.74 / 2.14 			& 1.39 / 1.68 / 2.24 	& \textbf{1.24 / 1.62 / 2.11}	& 20.8 (6.6 / 14.2) 		& 11.0 (8.6 / 2.4)	 & \textbf{4.0 (2.4 / 1.6)} \\ \hline
	\end{tabular}
	\label{tab:multi_agent_stats}
	\vskip -0.1in
\end{table*}

\subsection{Hardware Experiment}

A GA3C-CADRL policy implemented on a ground robot demonstrates the algorithm's performance among pedestrians.
We designed a compact, low-cost ($<~\$1000$) sensing suite with sensors placed as to not limit the robot's cargo-carrying capability~(\cref{fig:robot_sensors}).
The sensors are a 2D Lidar (used for localization and obstacle detection), and 3 Intel Realsense R200 cameras (used for pedestrian classification and obstacle detection).
Pedestrian positions and velocities are estimated by clustering the 2D Lidar's scan~\cite{Campbell13_NIPS}, and clusters are labeled as pedestrians using a classifier~\cite{liu2016ssd} applied to the cameras' RGB images~\cite{miller_dynamic_2016}.
Further details are in~\cite{everett_robot_2017}.

The lack of 3D Lidar is noteworthy, as it reduces the sensing suite's pricetag by an order of magnitude, yet also increases the uncertainty in the robot's knowledge about the state of the environment, particularly due to a reduction in perception range and accuracy.
The robot is still able to safely navigate in many challenging scenarios.
Future work will involve further analysis of the robot in more complicated environments to quantify the change in performance associated with the new algorithm and sensors.

A hardware video is included with this manuscript.

\section{Conclusion} \label{sec:conclusion}
This work presented a collision avoidance algorithm, GA3C-CADRL, that is trained in simulation with deep reinforcement learning without requiring any knowledge of other agents' dynamics.
It also proposed a strategy to enable the algorithm to select actions based on observations of an arbitrary number of nearby agents, using LSTM at the network's input.
The new approach is shown to outperform the existing method as the number of agents in the environment grows.
These results demonstrate the algorithm's ability to learn the problem's structure without it being explicity enforced, and support the use of LSTMs to encode a large number of agent states into a fixed-length representation of the world.
The new algorithm is also implemented on a small ground robot that is shown to navigate at human walking speed among pedestrians, without the use of a 3D Lidar.
Future work will leverage this paper's new, more general formulation to study the effects of signaling intent more explicitly through an agent's choice of action.





\section*{Acknowledgment}
This work is supported by Ford Motor Company.
The authors thank Shayegan Omidshafiei, Dong-Ki Kim, and NVIDIA for releasing open-source GA3C implementations.
\balance
\bibliographystyle{IEEEtran} 
\bibliography{biblio}

\begin{thebibliography}{10}
\providecommand{\url}[1]{#1}
\csname url@rmstyle\endcsname
\providecommand{\newblock}{\relax}
\providecommand{\bibinfo}[2]{#2}
\providecommand\BIBentrySTDinterwordspacing{\spaceskip=0pt\relax}
\providecommand\BIBentryALTinterwordstretchfactor{4}
\providecommand\BIBentryALTinterwordspacing{\spaceskip=\fontdimen2\font plus
\BIBentryALTinterwordstretchfactor\fontdimen3\font minus
  \fontdimen4\font\relax}
\providecommand\BIBforeignlanguage[2]{{%
\expandafter\ifx\csname l@#1\endcsname\relax
\typeout{** WARNING: IEEEtran.bst: No hyphenation pattern has been}%
\typeout{** loaded for the language `#1'. Using the pattern for}%
\typeout{** the default language instead.}%
\else
\language=\csname l@#1\endcsname
\fi
#2}}

\bibitem{van_den_berg_reciprocal_2008}
J.~{Van den Berg}, M.~Lin, and D.~Manocha, ``Reciprocal velocity obstacles for
  real-time multi-agent navigation,'' in \emph{Proceedings of the 2008 {{IEEE
  International Conference}} on {{Robotics}} and {{Automation}} ({{ICRA}})},
  2008, pp. 1928--1935.

\bibitem{chen_decentralized_2017}
Y.~Chen, M.~Liu, M.~Everett, and J.~P. How, ``Decentralized, non-communicating
  multiagent collision avoidance with deep reinforcement learning,'' in
  \emph{Proceedings of the 2017 {{IEEE International Conference}} on
  {{Robotics}} and {{Automation}} ({{ICRA}})}, Singapore, 2017.

\bibitem{Chen17_IROS}
Y.~F. Chen, M.~Everett, M.~Liu, and J.~P. How, ``Socially aware motion planning
  with deep reinforcement learning,'' in \emph{IEEE/RSJ International
  Conference on Intelligent Robots and Systems (IROS)}, Vancouver, BC, Canada,
  September 2017.

\bibitem{long2017deep}
P.~Long, W.~Liu, and J.~Pan, ``Deep-learned collision avoidance policy for
  distributed multiagent navigation,'' \emph{IEEE Robotics and Automation
  Letters}, vol.~2, no.~2, pp. 656--663, 2017.

\bibitem{long2017towards}
P.~Long, T.~Fan, X.~Liao, W.~Liu, H.~Zhang, and J.~Pan, ``Towards optimally
  decentralized multi-robot collision avoidance via deep reinforcement
  learning,'' \emph{arXiv preprint arXiv:1709.10082}, 2017.

\bibitem{li2017role}
M.~Li, R.~Jiang, S.~S. Ge, and T.~H. Lee, ``Role playing learning for socially
  concomitant mobile robot navigation,'' \emph{arXiv preprint
  arXiv:1705.10092}, 2017.

\bibitem{qi2018intent}
S.~Qi and S.-C. Zhu, ``Intent-aware multi-agent reinforcement learning,''
  \emph{Proceedings of the 2018 {{IEEE International Conference}} on
  {{Robotics}} and {{Automation}} ({{ICRA}}) (submitted)}, 2018.

\bibitem{sutskever2014sequence}
I.~Sutskever, O.~Vinyals, and Q.~V. Le, ``Sequence to sequence learning with
  neural networks,'' in \emph{Advances in neural information processing
  systems}, 2014, pp. 3104--3112.

\bibitem{cho2014learning}
K.~Cho, B.~Van~Merri{\"e}nboer, C.~Gulcehre, D.~Bahdanau, F.~Bougares,
  H.~Schwenk, and Y.~Bengio, ``Learning phrase representations using rnn
  encoder-decoder for statistical machine translation,'' \emph{arXiv preprint
  arXiv:1406.1078}, 2014.

\bibitem{hochreiter1997long}
S.~Hochreiter and J.~Schmidhuber, ``Long short-term memory,'' \emph{Neural
  computation}, vol.~9, no.~8, pp. 1735--1780, 1997.

\bibitem{tai2017virtual}
L.~Tai, G.~Paolo, and M.~Liu, ``Virtual-to-real deep reinforcement learning:
  Continuous control of mobile robots for mapless navigation,'' in
  \emph{Intelligent Robots and Systems (IROS), 2017 IEEE/RSJ International
  Conference on}.\hskip 1em plus 0.5em minus 0.4em\relax IEEE, 2017, pp.
  31--36.

\bibitem{tai2017socially}
L.~Tai, J.~Zhang, M.~Liu, and W.~Burgard, ``Socially-compliant navigation
  through raw depth inputs with generative adversarial imitation learning,''
  \emph{arXiv preprint arXiv:1710.02543}, 2017.

\bibitem{gupta2017cognitive}
S.~Gupta, J.~Davidson, S.~Levine, R.~Sukthankar, and J.~Malik, ``Cognitive
  mapping and planning for visual navigation,'' \emph{arXiv preprint
  arXiv:1702.03920}, vol.~3, 2017.

\bibitem{mnih2016asynchronous}
V.~Mnih, A.~P. Badia, M.~Mirza, A.~Graves, T.~Lillicrap, T.~Harley, D.~Silver,
  and K.~Kavukcuoglu, ``Asynchronous methods for deep reinforcement learning,''
  in \emph{International Conference on Machine Learning}, 2016, pp. 1928--1937.

\bibitem{babaeizadeh2017ga3c}
M.~Babaeizadeh, I.~Frosio, S.~Tyree, J.~Clemons, and J.~Kautz, ``Reinforcement
  learning thorugh asynchronous advantage actor-critic on a gpu,'' in
  \emph{ICLR}, 2017.

\bibitem{omidshafiei2017crossmodal}
S.~Omidshafiei, D.-K. Kim, J.~Pazis, and J.~P. How, ``Crossmodal attentive
  skill learner,'' \emph{arXiv preprint arXiv:1711.10314}, 2017.

\bibitem{alahi2016social}
A.~Alahi, K.~Goel, V.~Ramanathan, A.~Robicquet, L.~Fei-Fei, and S.~Savarese,
  ``Social lstm: Human trajectory prediction in crowded spaces,'' in
  \emph{Proceedings of the IEEE Conference on Computer Vision and Pattern
  Recognition}, 2016, pp. 961--971.

\bibitem{abadi2016tensorflow}
M.~Abadi, P.~Barham, J.~Chen, Z.~Chen, A.~Davis, J.~Dean, M.~Devin,
  S.~Ghemawat, G.~Irving, M.~Isard, \emph{et~al.}, ``Tensorflow: A system for
  large-scale machine learning.'' in \emph{OSDI}, vol.~16, 2016, pp. 265--283.

\bibitem{kingma2014adam}
D.~P. Kingma and J.~Ba, ``Adam: A method for stochastic optimization,''
  \emph{arXiv preprint arXiv:1412.6980}, 2014.

\bibitem{Campbell13_NIPS}
T.~Campbell, M.~Liu, B.~Kulis, J.~P. How, and L.~Carin, ``Dynamic clustering
  via asymptotics of the dependent dirichlet process mixture,'' in
  \emph{Advances in Neural Information Processing Systems 26}, 2013.

\bibitem{liu2016ssd}
W.~Liu, D.~Anguelov, D.~Erhan, C.~Szegedy, S.~Reed, C.-Y. Fu, and A.~C. Berg,
  ``Ssd: Single shot multibox detector,'' in \emph{European conference on
  computer vision}.\hskip 1em plus 0.5em minus 0.4em\relax Springer, 2016, pp.
  21--37.

\bibitem{miller_dynamic_2016}
J.~Miller, A.~Hasfura, S.~Y. Liu, and J.~P. How, ``Dynamic arrival rate
  estimation for campus {{Mobility On Demand}} network graphs,'' in \emph{2016
  {{IEEE}}/{{RSJ International Conference}} on {{Intelligent Robots}} and
  {{Systems}} ({{IROS}})}, Oct. 2016, pp. 2285--2292.

\bibitem{everett_robot_2017}
M.~Everett, ``Robot designed for socially acceptable navigation,'' Master
  Thesis, MIT, Cambridge, MA, USA, June 2017.

\end{thebibliography}
\end{document}